\documentclass[10pt, a4paper]{article}

\usepackage[]{lrec-coling2024} 

\usepackage{color}
\usepackage{lipsum}

\def\parsum#1{\bgroup \textcolor{blue}{Paragraph summary: #1}\egroup}
\def\sectionsum#1{\bgroup \textcolor{green}{Section content: #1}\egroup \\}

\usepackage{booktabs}

\usepackage{graphicx}

\usepackage{subcaption}

\newcommand{\prolific}{\textit{Prolific}}
\newcommand{\sosci}{\textit{SoSci Survey}}
\newcommand{\pavlovia}{\textit{Pavlovia}}
\newcommand{\wus}{\textit{Web-Use Skill}}
\newcommand{\iuipc}{\textit{IUIPC}}
\newcommand{\psychopy}{\textit{PsychoPy}}

\usepackage{longtable}

\title{To Share or Not to Share: What Risks Would Laypeople Accept to Give Sensitive Data to Differentially-Private NLP Systems?}

\name{Christopher Weiss$^1$ $\qquad$ Frauke Kreuter$^2$ $\qquad$ Ivan Habernal$^3$}

\address{
	$^1$ Trustworthy Human Language Technologies, \\ Department of Computer Science, Technical University of Darmstadt \\ 
	$^2$Department of Statistics, Ludwig-Maximilians-Universität München, Germany \\
	$^3$ Trustworthy Human Language Technologies, \\ Department of Computer Science, Paderborn University \\ 
	\texttt{www.trusthlt.org}
}

\abstract{
Although the NLP community has adopted central differential privacy as a go-to framework for privacy-preserving model training or data sharing, the choice and interpretation of the key parameter, privacy budget $\varepsilon$ that governs the strength of privacy protection, remains largely arbitrary. We argue that determining the $\varepsilon$ value should not be solely in the hands of researchers or system developers, but must also take into account the actual people who share their potentially sensitive data. In other words: Would \emph{you} share your instant messages for $\varepsilon$ of 10? We address this research gap by designing, implementing, and conducting a behavioral experiment (311 lay participants) to study the behavior of people in uncertain decision-making situations with respect to privacy-threatening situations. Framing the risk perception in terms of two realistic NLP scenarios and using a vignette behavioral study help us determine what $\varepsilon$ thresholds would lead lay people to be willing to share sensitive textual data -- to our knowledge, the first study of its kind.
\\ \newline  }

\begin{document}
	
\maketitleabstract

\section{Introduction}

The utilization of sensitive data in natural language processing (NLP) systems has become increasingly prevalent in recent years. Differential privacy is a widely-used method for privacy protection in training NLP models or publishing data \citep{igamberdiev-habernal-2022-privacy,igamberdiev-etal-2024-dp,senge-etal-2022-one,yin-habernal-2022-privacy,hu-etal-2024-differentially}.
However, current research on differential privacy in NLP mainly focuses on technical aspects, neglecting the human perception of the privacy risks.

The privacy risk of a differentially private algorithm is parametrized by $\varepsilon > 0$, the privacy budget.\footnote{Our focus is `pure' differential privacy. We leave exploring other popular flavors, such as $(\varepsilon, \delta)$-DP, or Rényi-DP for future work.}
While existing works consider appropriate $\varepsilon$ values  \citep{Bullek_Garboski2017,Xiong_Wang2020,Lee_Clifton_2011_How,Mehner_Voigt_Tschorsch_2021_Towards,Cummings_Kaptchuk_Redmiles_2021_I}, we still don't know how lay users perceive privacy risks and which $\varepsilon$ values are acceptable in which situations. In other words, for which $\varepsilon$ would \emph{you} give us your textual data?

Differential privacy makes no assumptions whether or not humans perceive the risk of privacy loss in the same way \citep{Dwork_McSherry2006}. However, studies on human risk assessment show that the perception of risks, especially when conveyed as probability, is dependent on various aspects and differs among humans \citep{Slovic_Peters_2006_Risk}. In this paper we thus ask the following research question. 
\textbf{What is the optimal value of $\varepsilon$ that lay users would accept and thus donate their sensitive text data?}

We systematically investigate this question by conducting a two-part survey study. First, we measure participants' attitudes towards privacy as well as their web-use skills, as \citet{Xiong_Wang2020,Cummings_Kaptchuk_Redmiles_2021_I} show that privacy concerns are an important factor when making decisions in privacy-threatening situations.
Second, we design, develop, and conduct a behavioral experiment which involves repeated risk assessments in privacy-threatening situations from our participants.
We frame the risks in two prototypical NLP scenarios, namely the hypothetical collection of text medical records and collecting a chat history of an instant messenger.
The primary objective of collecting data with our survey and the behavioral experiment is to gain insights into our research question, with the aim of confirming our hypothesis that human decision-making behavior can be effectively modeled by a logistic function within the domain under investigation.

\section{Related work}
\label{ch:03_Related_Work}

The theoretical background on differential privacy and the measurement of local risks including several examples is detailed in Appendix~\ref{ch:02_Theoretical_Background}.


\citet{Bullek_Garboski2017} conducted a study on how users understand privacy parameters in randomized response. Their findings show that if end-users understand the privacy preserving data perturbations, they are more likely feel trust and comfort when sharing data.
However, their study investigates a local differential privacy technique, not a global differential privacy environment.

\citet{Xiong_Wang2020} examined how informing users that their information is protected with different differential privacy techniques influences their willingness to share low and high sensitive information.
As opposed to \citet{Bullek_Garboski2017}, their study encompasses both global and local differential privacy approaches.
However, the study does not address the determinacy of specific epsilon values in meeting the privacy requirements of participants, nor does it examine the existence of a threshold for decision-making.

\citet{Lee_Clifton_2011_How} on the other hand claim that the parameters of differential privacy have an intuitive theoretical interpretation, but choosing appropriate values is non-trivial.
Based on a series of theoretical computations they show that $\varepsilon=0.3829$ would be an appropriate value. However, their computation solely relies on the mathematical foundations of differential privacy and its parameters. The calculations of \citet{Lee_Clifton_2011_How} completely disregard human perception, the need for privacy, and the additional factors already mentioned that influence the decision to share data.

The research of \citet{Mehner_Voigt_Tschorsch_2021_Towards} is based on the findings of \citet{Lee_Clifton_2011_How}. \citet{Mehner_Voigt_Tschorsch_2021_Towards} put the $\varepsilon$ value as privacy loss parameter of differential privacy in the center of their work.
Given the lack of understanding of the privacy guarantees $\varepsilon$, they provide more understandable statements on the privacy loss and introduce the notion of a global privacy risk, global privacy leak and local privacy leak as comprehensible measures to communicate privacy loss to end users, yet they remain inconclusive about actual $\varepsilon$ values.
\citet{Cummings_Kaptchuk_Redmiles_2021_I} studied differential privacy from the user's perspective, focusing on how users’ privacy expectations relate to differential privacy as they are likely to encounter it in-the-wild.
While \citet{Cummings_Kaptchuk_Redmiles_2021_I} supports the significance of considering participants' privacy concerns, their study primarily concentrates on end-users' comprehension of differential privacy and does not address the varying degrees of privacy based on the $\varepsilon$ value.


Our paper fills the research gap. As opposed to \citet{Bullek_Garboski2017}, we conduct research in the environment of global differential privacy.
We incorporate the work of \citet{Lee_Clifton_2011_How} which proposed a purely mathematical way to determine appropriate epsilon values, but negate human perception, the need for privacy, and the additional factors, i.e., the privacy concerns. The work of \citet{Mehner_Voigt_Tschorsch_2021_Towards}, which is based on the work of \citet{Lee_Clifton_2011_How}, provides a worst-case, but more comprehensible notion to communicate privacy loss to end-users. We follow their suggestion to conduct a user study using their privacy risk notion called local privacy leak converted to natural frequencies. We present these risks in a behavioral experiment that follows an adapted form of the vignette design used by \citet{Cummings_Kaptchuk_Redmiles_2021_I}, which enables our participants to relate to the context and the decision they will be making.

\label{ch:04_Methods}

\section{Methodology}
\label{sec:04_Technical_Equipment}

\subsection{Study setup}

Our online study consists of two parts: a survey, which focuses on measuring privacy attitudes using the \iuipc\ and the \wus\ questionnaires and a behavioral experiment measuring privacy risk assessment in the form of a vignette design.

We used \prolific\ as a participant recruitment tool, combined with \sosci\footnote{\url{https://www.soscisurvey.de/}} service which hosted the developed questionnaire.
The behavioral experiment was developed using \psychopy\ version 2022.2.3 and hosted on \href{https://pavlovia.org/}{\pavlovia} \citep{Peirce_Gray_2019_PsychoPy2}.
The median completion time was $09m\,55s$. At the end of the experiment, participants were automatically redirected to \prolific, where the platform asked them if there were any problems and if they would like to contact the responsible researcher of the study. If this was not the case, the study was considered successfully completed and the participants got \textsterling 1.50 for their participation. The overall budget for this study was $\sim$ 800 €.



\subsection{Participants}
\label{sec:04_Participants}



The paid service \prolific\ was used to recruit participants for the survey and experiment. We applied the following pre-selection criteria: (a) living in USA, Canada, UK, Germany, Austria, and Switzerland, (b) fluency in English, and (c)  approval rate of at least 95\% and had to have participated in at least 100 studies on \prolific.

The entire study, comprised of the survey and our behavioral experiment, was conducted online using the \prolific\ participant recruitment service in November 2022. \prolific\ provides researchers with a certain amount of demographic data about participants. In addition to the data gathered by the pre-screening algorithms applied, \prolific\ provides data on 16 variables per participant, protected by privacy regulations of \prolific.
After removing several participants due to technical failures, we ended up with $311$ participants whose data is used for analysis.

Of the $311$ participants, $155$ identified themselves as women ($49.84\%$) and $156$ as men ($50.16\%$). Participants have a mean age of $42.83$ years (median: $41.0$ years) with a standard deviation of $14.39$ years. On average, female participants are $41.25$ years old with a standard deviation of $13.07$ years whereas male participants are $44.39$ years old with a standard deviation of $15.46$ years. Of the total $311$ participants, $87.14\%$ resided in the United Kingdom, $4.82\%$ in Germany, $4.5\%$ in Canada, and the remaining $3.54\%$ were distributed among Switzerland, the United States, and Austria at the time of conducting the study.

All $311$ participants reported fluency in English. Furthermore, $85.45\%$ reported not currently being an enrolled student, whereas, the remaining $14.55\%$ of participants reported being enrolled students. $44.0\%$ of participants are full-time employees, whereas $24.8\%$ are not in paid work, $22.8\%$ work part-time, and the remaining $8.4\%$ are either unemployed, starting a new job within the next month, or have selected the ``\textit{Other}'' option. $34.41\%$ of participants reported a high-school degree as their highest level of education. Whereas $56.59\%$ hold a university degree and $2.57\%$ reported holding a PhD or higher value degree. The remaining $6.43\%$ chose either the ``\textit{Other}'' or the ``\textit{Prefer not to say}'' option, respectively.

This section shows that the participant sample offers considerable diversity in terms of the variables gender, age, employment status and highest educational degree, which is an advantage in terms of generalizability of the results. The skewnesses in country of residence can potentially lead to the sample bias as the majority of respondents reside in the United Kingdom. This fact may affect the generalizability of the results and should be taken into account when interpreting them.



\subsection{Questionnaire}
\label{sec:04_Questionnaire}
The questionnaire using \sosci\ represents the first part of our study. The aim of the survey was to measure the basic attitude towards privacy in technical systems. For this purpose, we used the already created and validated questionnaire on the construct ``\textit{Internet Users' Information Privacy Concerns (IUIPC)}'' by \citet{Malhotra_Kim_Agarwal_2004_Internet}. We will use the characteristics of the participants with respect to this construct as a baseline or potential scaling factor when comparing data collected during our behavioral experiment among different individual participants or groups of participants.

It is straightforward to comprehend that different baseline attitudes toward privacy have implications for the respective risk estimation in privacy-threatening situations. For instance, if privacy is assessed as fundamentally not that important (accompanied by a low \iuipc score), this could be reflected in a more relaxed risk assessment and vice versa. The question that arises based on this train of thought is the following: Does personal attitude towards privacy have a significant impact on risk assessment, so that instead of finding one optimal epsilon value, several corresponding values must be determined for different groups based on their \iuipc scores?

Another concept to which our questionnaire measures the respective expressiveness in participants is the so-called \wus, which was developed by \citet{Hargittai_Hsieh_2012_Succinct}. The scope of this work is on natural language processing systems. Considering how end\=/users usually interact with such systems, it becomes apparent that they are mainly used via interfaces on the Internet. For this reason it makes sense to determine the proficiency of the participants with the Internet.

It should be noted that the \wus\ score is calculated based on the self-assessment of the participants. However, since the \wus\ score is only to be used as a basis for dividing the participants into groups and does not represent the dependent variable to be investigated, the bias regarding systematic under- or overestimation of the skills queried by \wus\ items was not controlled for as a confounding variable.

\section{Behavioral experiment}
\label{sec:04_Experiment}

We aim to complement the existing knowledge on differential privacy \citep{Dwork_McSherry2006, Cummings_Kaptchuk_Redmiles_2021_I,Wood_Altman_2018_Differential,Xiong_Wang2020,Bullek_Garboski2017} with insights from the research fields of psychology and cognitive science. Our experiment is designed to gather initial data on human behavior in privacy-related risk situations and provide systematic insights into human risk perception.

\subsection{Scenarios of sensitive text data sharing}

The experiment had two instances of the independent variable \textit{Scenario}. The first scenario (\textit{medical}) is adapted from \citet{Cummings_Kaptchuk_Redmiles_2021_I}. This scenario describes a situation in which the participant is asked by the primary care physician whether their medical record can be shared with a non-profit organization to help medical research improve treatment methods via building automatic NLP systems.

The second scenario (\textit{language}) is created according to a similar template in order to have a direct relation to NLP. In this scenario, participants are asked to imagine that a non-profit organization wants to build an app that will allow anyone in the world to learn a new language for free, using the participants' instant messenger conversations from the last 30 days.

The motivation for these scenarios is twofold. First, they differ in the domain they address. This work is concerned with participants' risk perception and privacy attitudes in the context of NLP. For this reason, one scenario is located in this specific domain and the other is not. Second, literature shows people tend to perceive medical data as more worthy of protection than their messenger data \citep{Xiong_Wang2020}. The participants were randomly assigned to one of the two scenarios. Among the $311$ participants, $147$ were randomly assigned to the \textit{medical} scenario and $164$ to the \textit{language} scenario. Accordingly, each participant was exposed to only one scenario.

The vignette-based design was used to elicit respondents intended behavior, as such studies have been found to well-approximate real-world behavior \citep{Hainmueller_Hangartner_Yamamoto_2015_Validating}.
In both scenarios the privacy of data would be secured by applying differential privacy. The fact that the data were protected by differential privacy was not disclosed to the participants in order to avoid bias due to different levels of knowledge or confusion. For the task, it is only relevant that the participants understand what data is involved and the risks regarding re-identification, which is more generally described as misuse of the data.


\subsection{Experiment design}
\label{subsec:04_Experiment_Design}
The behavioral experiment is a \textit{yes-no} task and we implemented it as a 2~(\textit{Scenario})~x 5~(\textit{Amount of Data Subjects})~x~9~($\varepsilon$-\textit{Value}) between-subject design. This design results in three independent variables: \textit{Scenario}, \textit{Amount of Data Subjects} and $\varepsilon$-\textit{Value}.


The quantification of risk we used in this study is rooted in the work of \citet{Mehner_Voigt_Tschorsch_2021_Towards} and serves to convert the abstract concept of an epsilon value as privacy loss parameter into a human-comprehensible representation of risk called the \textit{local privacy leak} introduced by \citet{Mehner_Voigt_Tschorsch_2021_Towards}. This metric is scaled between $0$ and $1$ and can be interpreted as probability. Several studies have shown that it is easier for humans to understand and assess risks described using \textit{natural frequencies} ($\omega$) instead of plain probabilities \citep{Gigerenzer_2011_What, Hoffrage_Gigerenzer_2002_Representation, Mehner_Voigt_Tschorsch_2021_Towards}. Assume there is a probability $\Pr = 0.01$ that data will be misused in any kind of way. Instead of telling users ``\textit{There is a 1\% chance of data misuse}'', one should re-formulate the risk using the pattern: ``\textit{In 1 out of 100 cases data misuse can occur}''.

The \textit{local privacy leak} is computed based on the privacy loss parameter $\varepsilon$, which represents the eponymous independent variable $\varepsilon$-\textit{Value} as well as on number of data subjects $n$, which depicts the independent variable \textit{Amount of Data Subjects}. Therefore, we systematically changed these independent variables in the experiment to examine their influences on human risk assessment.

It is very important to distinguish between the independent variable \textit{Amount of Data Subjects} and the actual number of participants with respect to our study. The independent variable \textit{Amount of Data Subjects}, also represented by $n$, describes the number of distinct data subjects held by the trusted curator in the imaginary scenario described to the participants and thus has an impact on the risk represented by the \textit{local privacy leak}.


The behavioral experiment consisted of $225$ trials for each participant, with $45$ distinct combinations of the independent variables \textit{Amount of Data Subjects} and $\varepsilon$-\textit{Value} determining the level of risk for data misuse for each trial. To obtain meaningful data for each distinct combination of the two independent variables, we repeated each of the $45$ combinations five times.\footnote{This high number of repetitions is important for precise response estimates. Such repetitions are typically conducted in behavioral experiments.} To ensure that the results of the experiment were not influenced by order effects, we randomized the trials. Each participant was presented with a series of decision-making situations on a computer screen and was asked to indicate their willingness to share personal information as described by the given scenario description by pressing the right or left arrow key on their keyboard. The dependent variables in this study were the participant's response and the response time. The response which was either ``\textit{Share}'', to indicate they accept sharing their data in this trial or ``\textit{Don't share}''. The response time was measured in milliseconds and recorded for each trial.



\subsection{Experiment stimuli}
\label{subsec:04_Stimuli}
In the behavioral experiment, we used a set of stimuli to manipulate the independent variables \textit{Amount of Data Subjects} and $\varepsilon$-\textit{Value} as these variables have an impact on the risk of data misuse represented by the \textit{local privacy leak}.
The given risk is itself represented by using \textit{natural frequencies}: ``\textit{In $1$ out of $\omega$ cases data misuse can occur}'', where the value $\omega$ is used to manipulate the risk on a trial-by-trial basis.
Overall we had 45 different stimuli combinations.

\paragraph{Amount of Data Subjects} The independent variable \textit{Amount of Data Subjects} ($n$) describes the amount of different data subjects to which the respective scenario (\textit{medical} or \textit{language}) is referring to. They correspond to the size of the dataset held by the trusted curator who, for instance, wants to train a privacy-preserving model. We used $n \in$ (1,000; 10k; 100k; 1M; 10M).


\paragraph{Epsilon Value} The independent variable $\varepsilon$-\textit{Value} represents the privacy loss parameter of the concept of differential privacy.
While the NLP researchers are interested in accurate results of their analyses or models and thus prefer higher values of epsilon, the data subject often cannot assess the privacy risks of sharing their data. However, they tend to prefer lower epsilon values.
With an epsilon value of $0.01$, the subjects should clearly predominantly agree to sharing the data. Whereas an epsilon value of $10$ corresponds to a very high risk of data misuse and it is therefore expected that the majority of subjects will object to the sharing of the data. We used $\varepsilon \in $ (0.01; 0.1; 0.5; 1; 2; 3; 4; 5; 10).

\section{Results and analysis}
\label{ch:05_Analysis_Discussion}
\subsection{Descriptive analysis}
\label{subsec:05_DA_Behav_Exp_Desc_Ana}
\begin{figure}
	\centering
	\includegraphics[width=\linewidth]{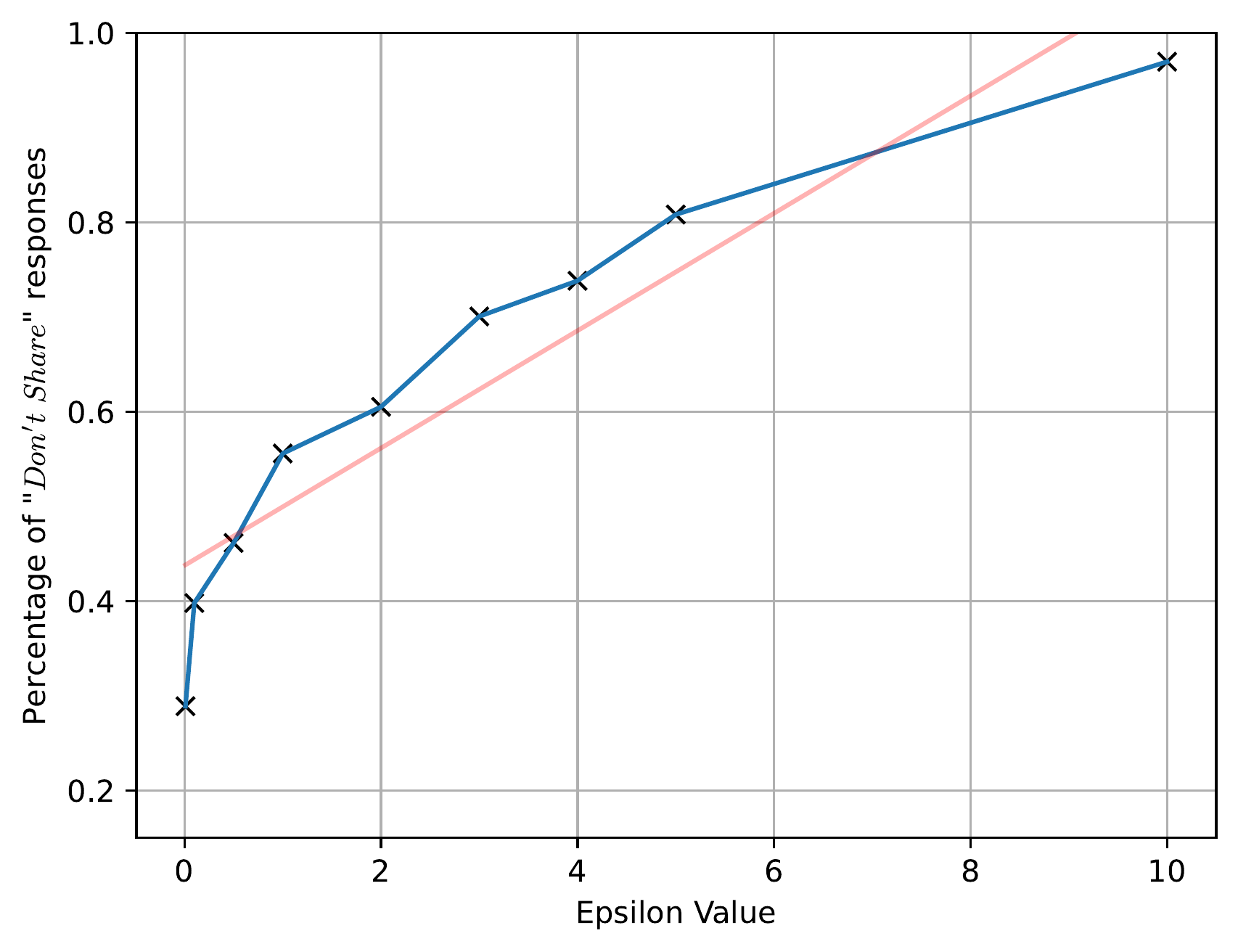}
	\caption{Percentage of ``\textit{Don't Share}'' responses per $\varepsilon$-value}
	\label{fig:05_DA_Behav_Exp_Perc_No_General}
\end{figure}

Figure \ref{fig:05_DA_Behav_Exp_Perc_No_General} shows the proportion of ``\textit{Don't Share}'' responses of all subjects per epsilon value. It is clear that the number of ``\textit{Don't Share}'' responses increases as the epsilon value increases. This is also backed up by the Pearson correlation $r=0.93$ which is a statistically significant positive correlation ($p=0.0003$).


\begin{figure}[ht]
	\centering
	\includegraphics[width=\linewidth]{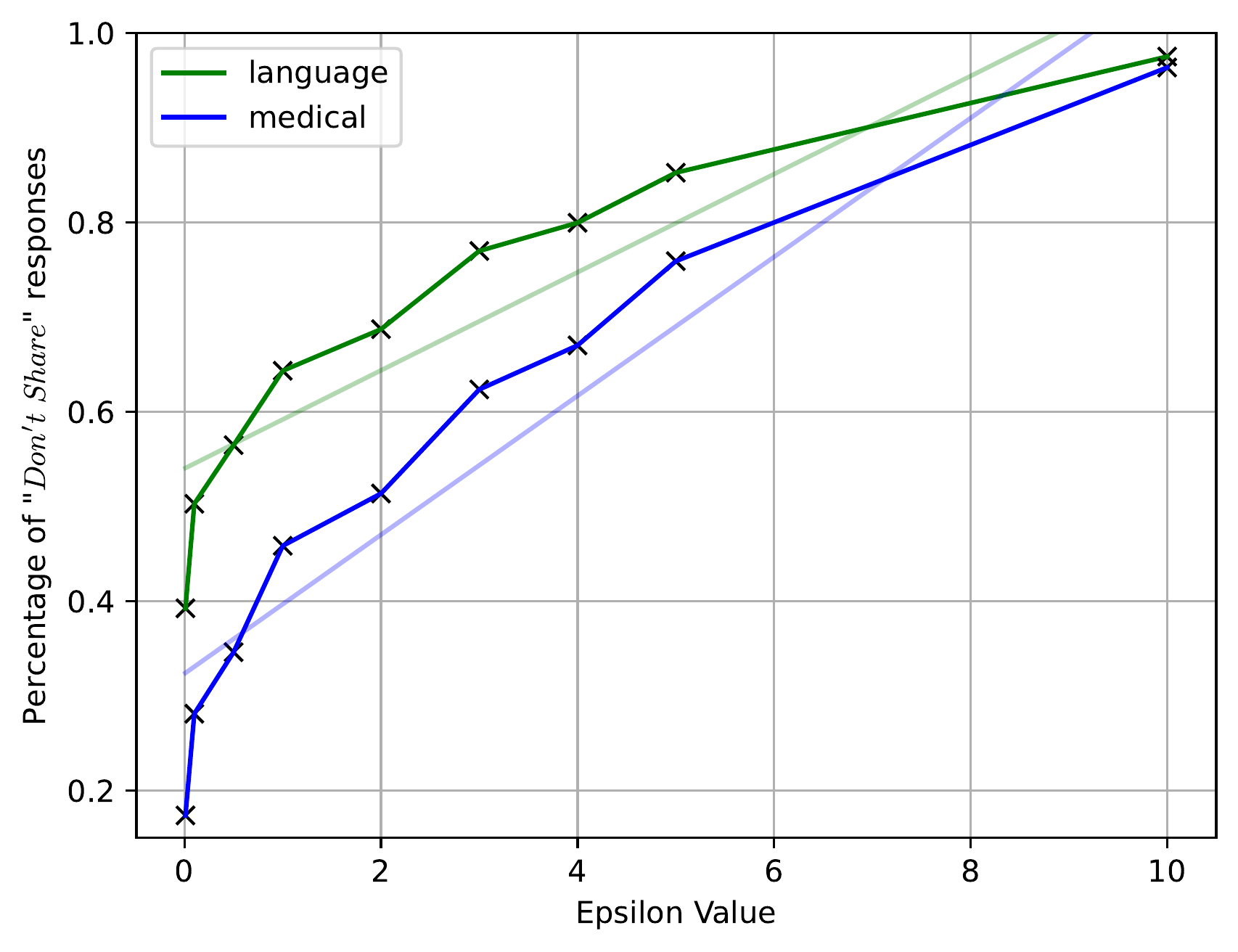}
	\caption{Percentage of ``\textit{Don't Share}'' responses per $\varepsilon$-value grouped by scenario}
	\label{fig:05_DA_Behav_Exp_Perc_No_General_per_Scenario}
\end{figure}

Breaking down the results into the two different scenarios (\textit{language} and \textit{medical}) results in Figure \ref{fig:05_DA_Behav_Exp_Perc_No_General_per_Scenario} which shows that participants who were assigned to the \textit{language}-scenario gave a higher proportion of ``\textit{Don't Share}'' responses, even at lower epsilon values, compared to the participants in the \textit{medical}-scenario. This suggests that, contrary to current opinion \citep{Xiong_Wang2020}, the data from the \textit{language}-scenario is considered more sensitive and worthy of protection than the data of the \textit{medical}-scenario.


\begin{figure}
	\centering
	\includegraphics[width=\linewidth]{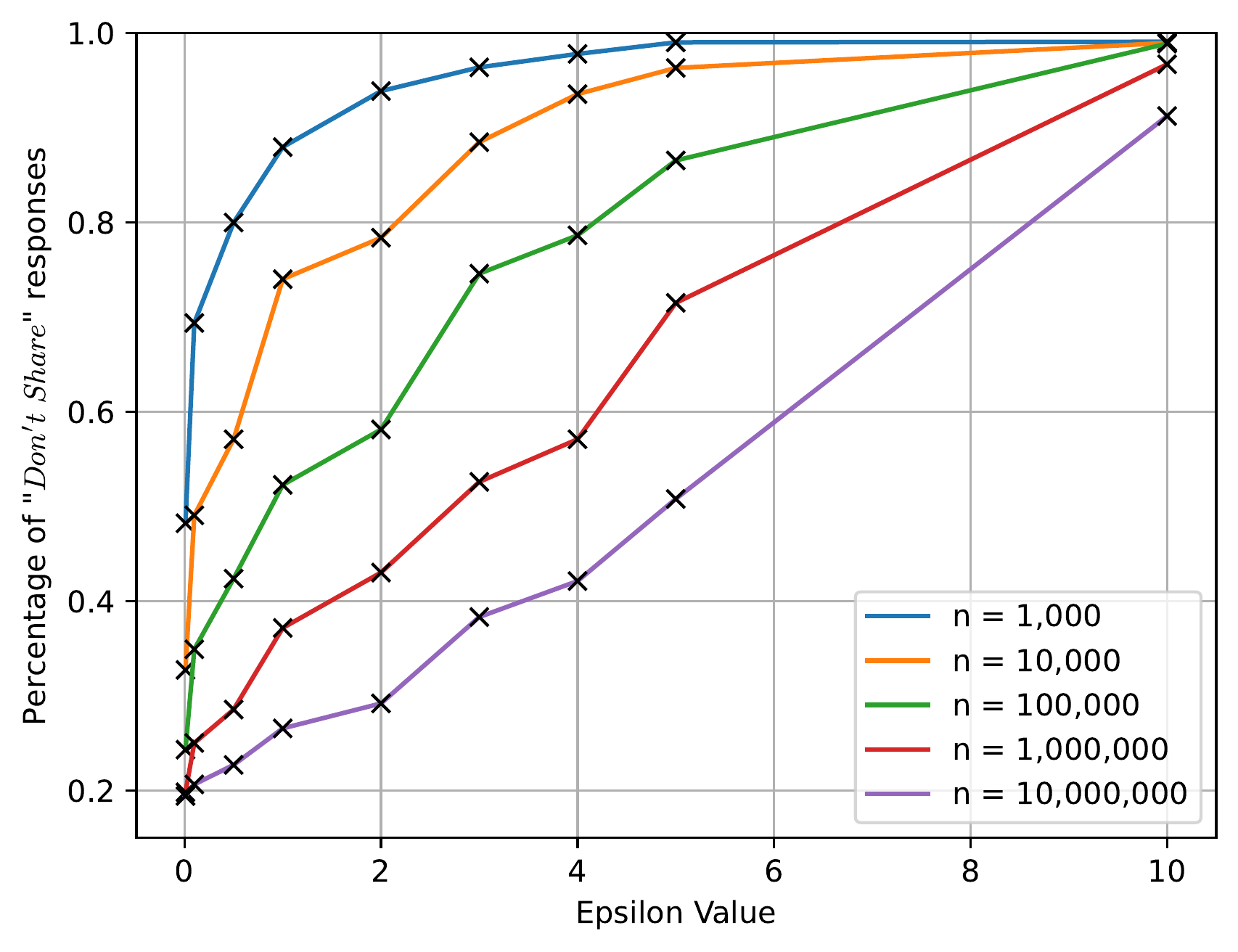}
	\caption{Percentage of ``\textit{Don't Share}'' responses per \textit{Amount of Data Subjects} ($n$) and $\varepsilon$-value}
	\label{fig:05_DA_Behav_Exp_Perc_No_per_n}
\end{figure}


Breaking down the results along the $n$ dimension (the hypothetical size of the resulting data set to be collected by the trusted curator), Figure~\ref{fig:05_DA_Behav_Exp_Perc_No_per_n} shows five lines, one for each level of the independent variable \textit{Amount of Data Subjects} ($n$). The smaller $n$, the steeper the curve and thus the number of ``\textit{Don't Share}'' responses increases.

\begin{figure}[ht]
	\centering
	\includegraphics[width=\linewidth]{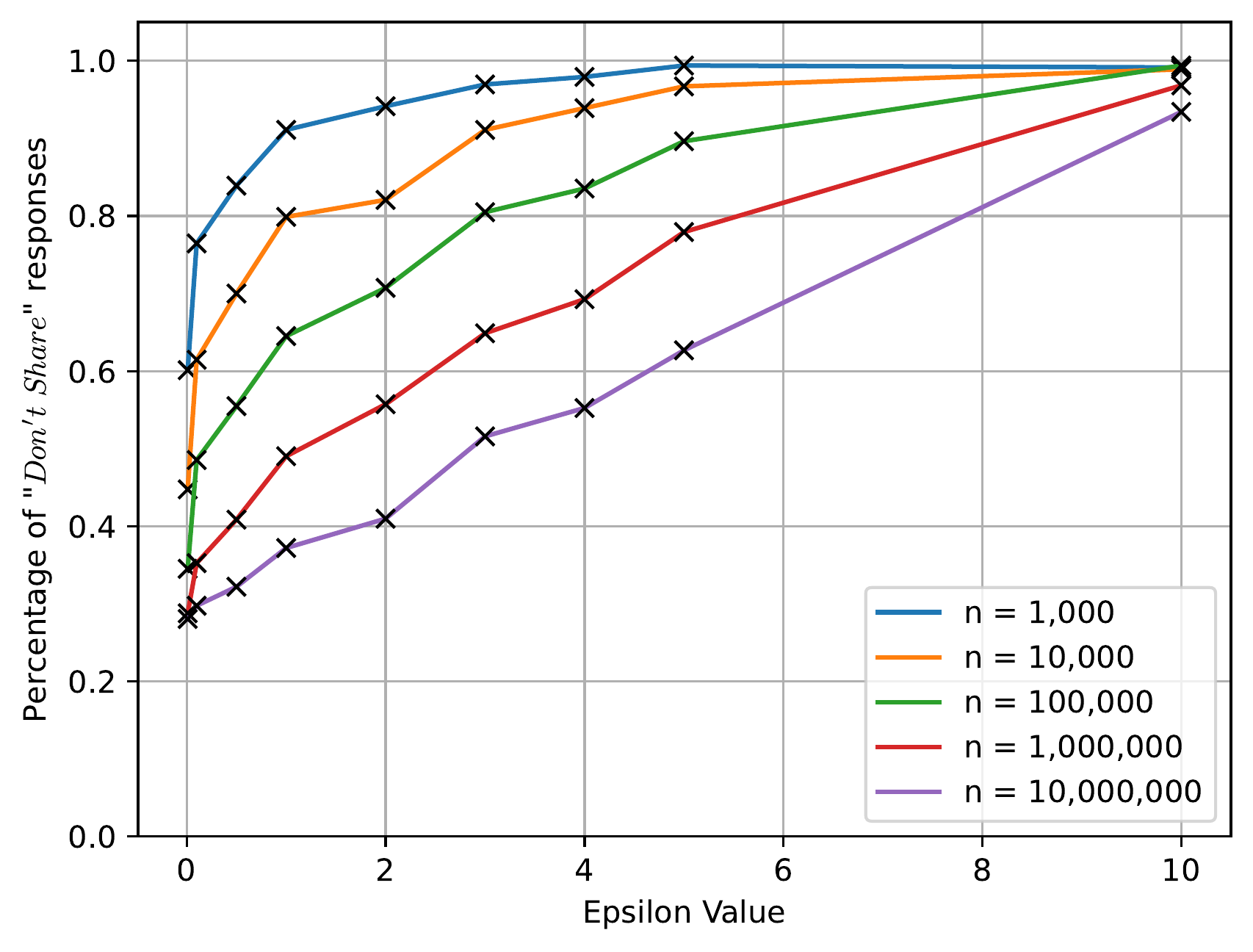}
	\caption{Percentage of ``\textit{Don't Share}'' responses per $n$ and $\varepsilon$ in the \textit{Language}-scenario}
	\label{fig:05_DA_Behav_Exp_Perc_No_per_n_language}
\end{figure}

\begin{figure}[ht]
	\centering
	\includegraphics[width=\linewidth]{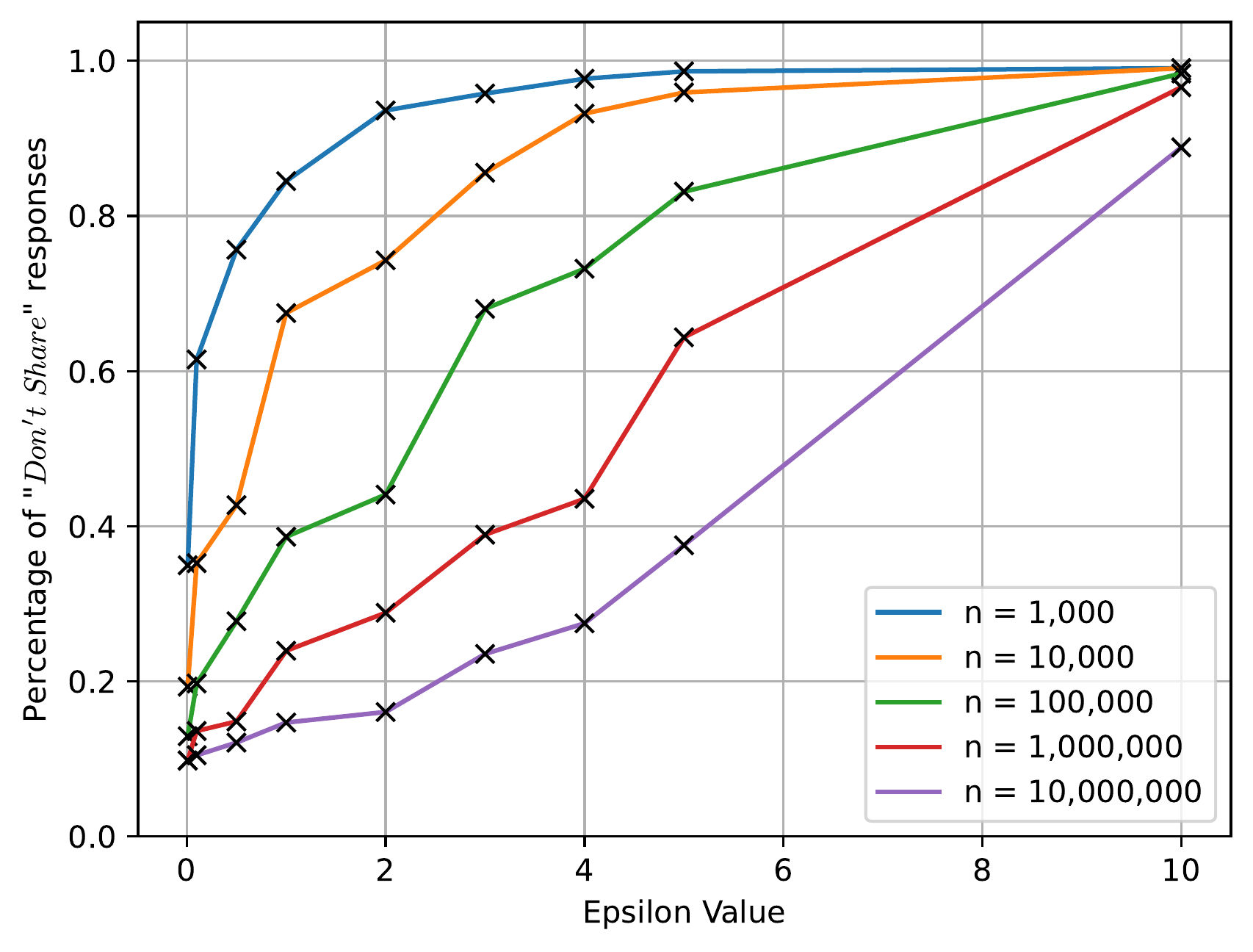}
	\caption{Percentage of ``\textit{Don't Share}'' responses per $n$ and $\varepsilon$ in the \textit{Medical}-scenario}
	\label{fig:05_DA_Behav_Exp_Perc_No_per_n_medical}
\end{figure}


Figures \ref{fig:05_DA_Behav_Exp_Perc_No_per_n_language} and \ref{fig:05_DA_Behav_Exp_Perc_No_per_n_medical} break down the two different scenarios (\textit{language} and \textit{medical}). They show
that not only the independent variable \textit{Amount of Data Subjects} has an influence on the decision-making behavior of the participants but the scenario (\textit{language} or \textit{medical}) also plays an important role.


\subsection{Psychometric functions and $\varepsilon$-thresholds}
\label{subsec:05_EA_Behav_Exp_Psych_Func_Thres}
The prior examinations have demonstrated that there is no straightforward resolution to our research question regarding a general and optimal epsilon value in relation to individuals' risk perception and decision-making behavior. The results have indicated that both the scenarios (\textit{language} and \textit{medical}) and the independent variable \textit{Amount of Data Subjects} have a significant impact on the participants' decision-making behavior.

Rather than determining a single general and optimal epsilon value, we can now utilize psychometric functions to determine epsilon thresholds as guidelines, taking into consideration the independent variables \textit{Scenario} and \textit{Amount of Data Subjects}. These guidelines may serve as a substitute for the previously sought after single optimal epsilon value. In this work we use the logistic function as  a psychometric function \citep{Kingdom_Prins_2016_Psychophysics}.

\begin{figure}[ht]
	\centering
	\includegraphics[width=\linewidth]{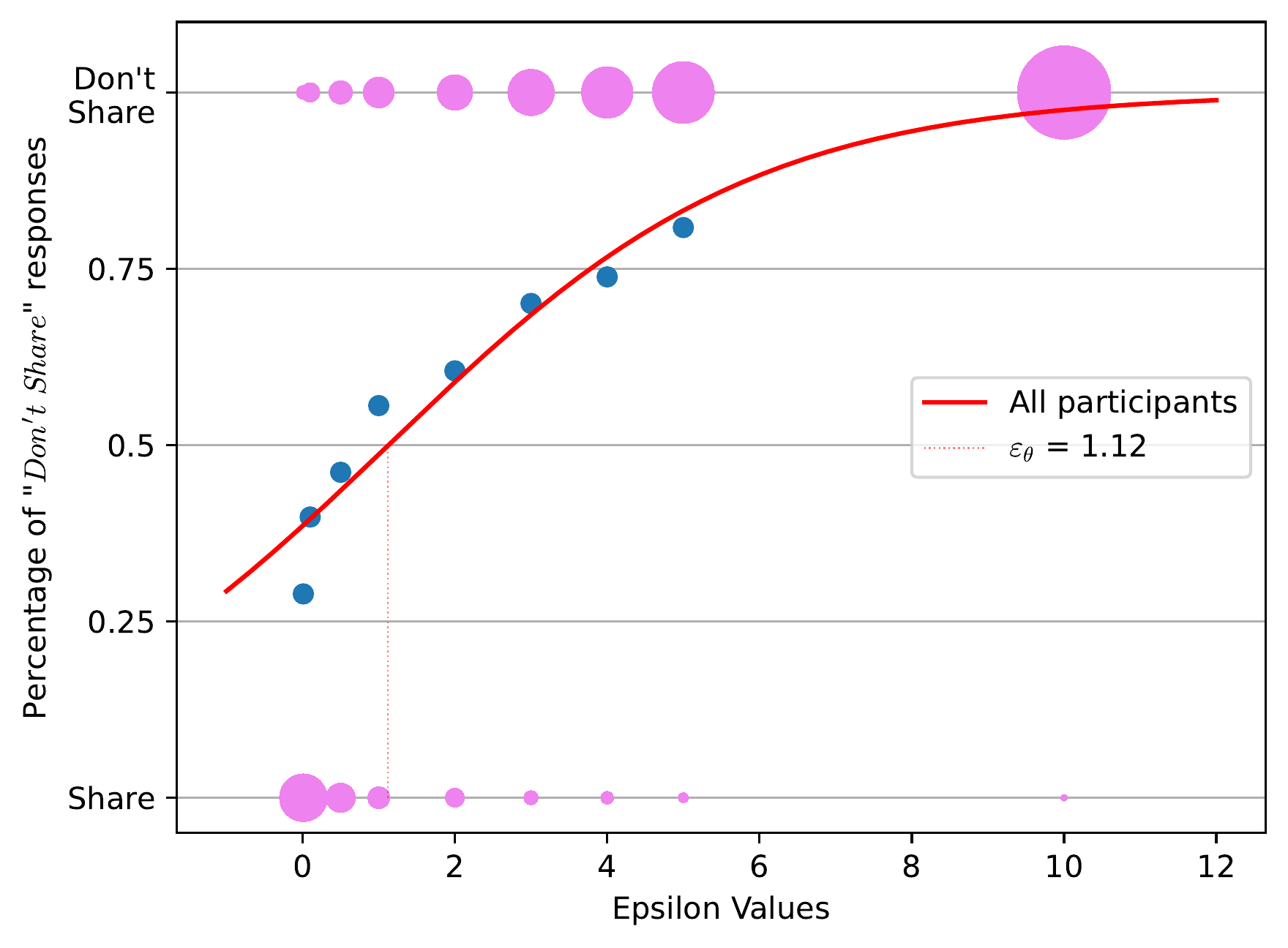}
	\caption{Psychometric curve over all participants and conditions}
	\label{fig:05_EA_Behav_Exp_General_Psych}
\end{figure}

Figure \ref{fig:05_EA_Behav_Exp_General_Psych} shows the percentage of ``\textit{Don't Share}'' responses for all participants across all conditions as blue circles. In violet are the original ``\textit{Share}'' or ``\textit{Don't Share}'' responses depicted that we used to fit the psychometric function. It is common for a $y$-value of $y=0.5$ to use the corresponding $x$-value of the fitted function as a threshold \citep{Kingdom_Prins_2016_Psychophysics}. In \ref{fig:05_EA_Behav_Exp_General_Psych}, the threshold is $\varepsilon_{\theta}=1.12$, which is represented by the vertical, dashed, red line. This implies that with an epsilon value of $\varepsilon_{\theta}=1.12$, the majority of people agree to share data. We determined the goodness of fit by evaluating the Root Mean Square Error $\mathit{RMSE}=0.04$ and R-squared $r^2=0.95$, which each indicated a very good fit.

\begin{figure}[ht]
	\centering
	\includegraphics[width=\linewidth]{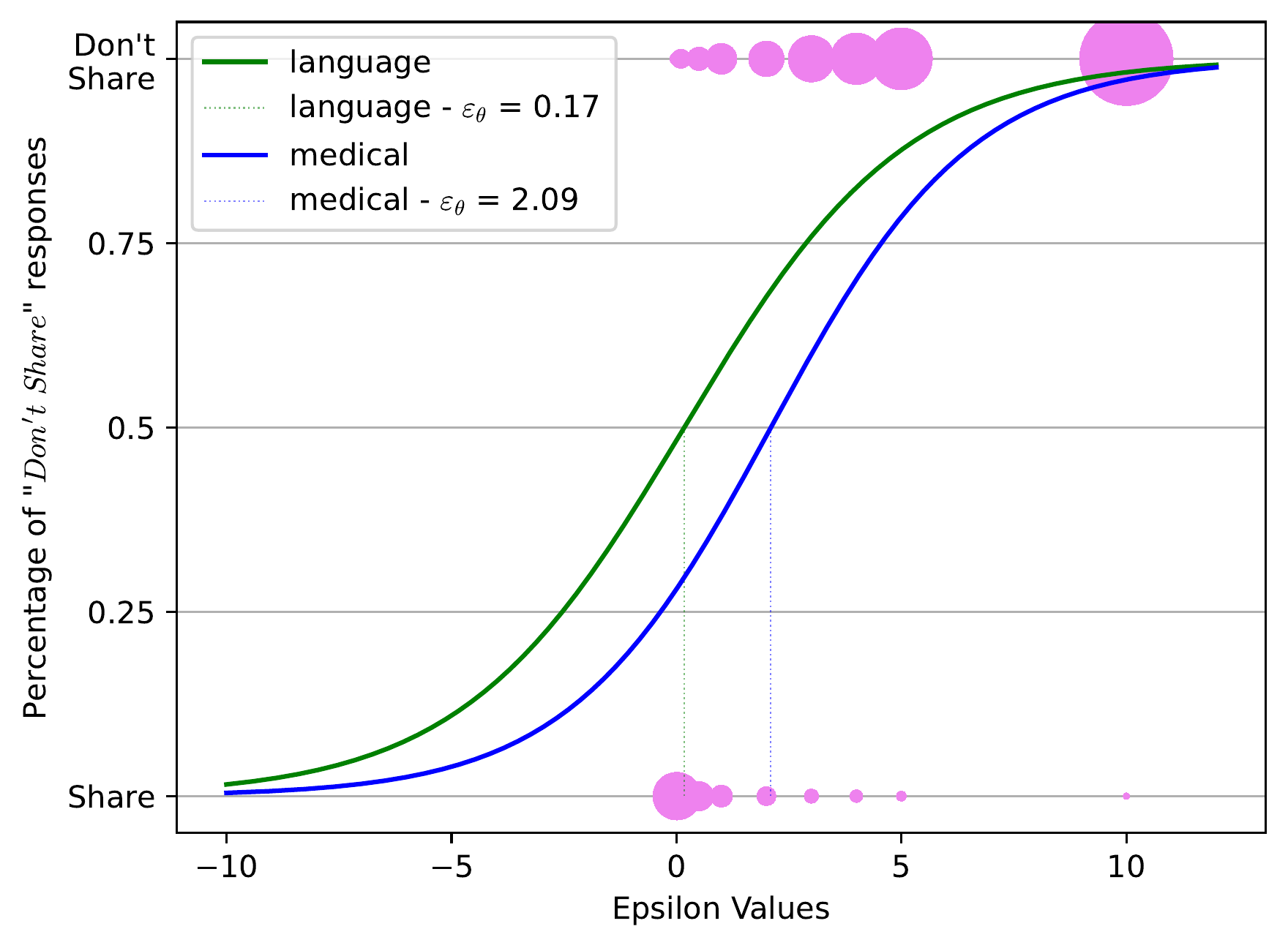}
	\caption{Psychometric curve over all participants separated by scenario}
	\label{fig:05_EA_Behav_Exp_General_Psych_per_Scenario}
\end{figure}

Given the observed differences in responses to the two scenarios, Figure \ref{fig:05_EA_Behav_Exp_General_Psych_per_Scenario} depicts the psychometric curves fitted to the decision-making data of each scenario.
The threshold for the language scenario is at an epsilon value of $\varepsilon_{\theta \textrm{lang}}=0.17$, compared to the threshold for the medical scenario of $\varepsilon_{\theta  \textrm{med}}=2.09$. This finding reinforces the previous assumption that the participants in our experiment consider the data from the language scenario to be more privacy-sensitive than the data from the medical scenario.

\paragraph{Key take-aways}

\begin{table}[h!]
	\centering
	\begin{tabular}{rrrr}
		\toprule
		$n$ & $\varepsilon_{\theta  \textrm{both }}$ & $\varepsilon_{\theta  \textrm{lang}}$ & $\varepsilon_{\theta \textrm{med}}$ \\ \midrule
		All $n$ values   & 1.12    & 0.17    & 2.09 \\ \midrule
		1,000          & 0.00   & 0.00   & 0.08 \\
		10,000         & 0.31    & 0.00   & 0.80 \\
		100,000        & 1.29    & 0.38    & 2.18 \\
		1,000,000       & 2.81    & 1.58    & 4.02 \\
		10,000,000      & 4.54    & 3.01    & 5.93 \\ \bottomrule 
	\end{tabular}
	\caption{Epsilon threshold results. Zero $\varepsilon$ values mean that the majority of participants would not share data for this particular scenario and $n$.}
	\label{tab:06_Epsilon_Thres}
\end{table}

By distinguishing between the two independent variables \textit{Scenario} and \textit{Amount of Data Subjects}, it again becomes abundantly evident that, at least based on our data, there seems to be no one general and optimal epsilon value. It is important to consider the type of data and the underlying number of data sets.
The key findings and the respective epsilon thresholds are summarized in Table \ref{tab:06_Epsilon_Thres}.



\paragraph{Do \iuipc\ and \wus\ scores correspond to $\varepsilon$ thresholds?}

\label{subsec:05_EA_Behav_Exp_IUIPC_WUS_Psych_Func_Thres}
According to \citet{Xiong_Wang2020} personal privacy considerations hold a significant influence on the decision-making behavior of participants in relation to differential privacy. This was the reason we chose the \iuipc\ questionnaire developed by \citet{Malhotra_Kim_Agarwal_2004_Internet} to measure the privacy concerns of our participants. If the \iuipc\ scores are related to the epsilon thresholds, the \iuipc\ questionnaire could be used as a predicting variable.

\begin{figure}[htb]
	\centering
	\includegraphics[width=0.99\linewidth]{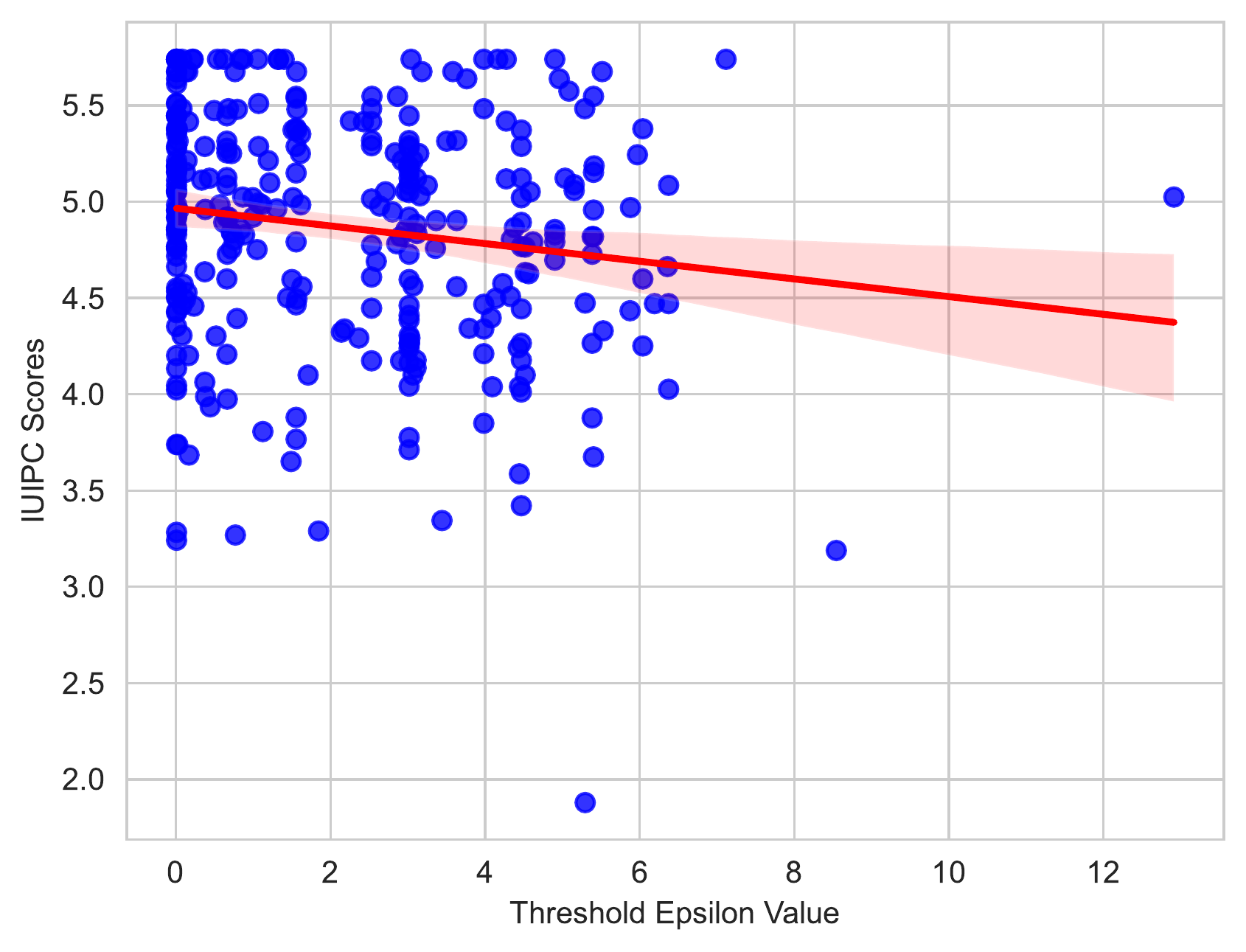}
	\caption{$\varepsilon$-Thresholds and \iuipc\ scores of all participants}
	\label{fig:05_EA_Behav_Exp_IUIPC_Thres}
\end{figure}

\begin{figure}[htb]
	\centering
	\includegraphics[width=0.8\linewidth]{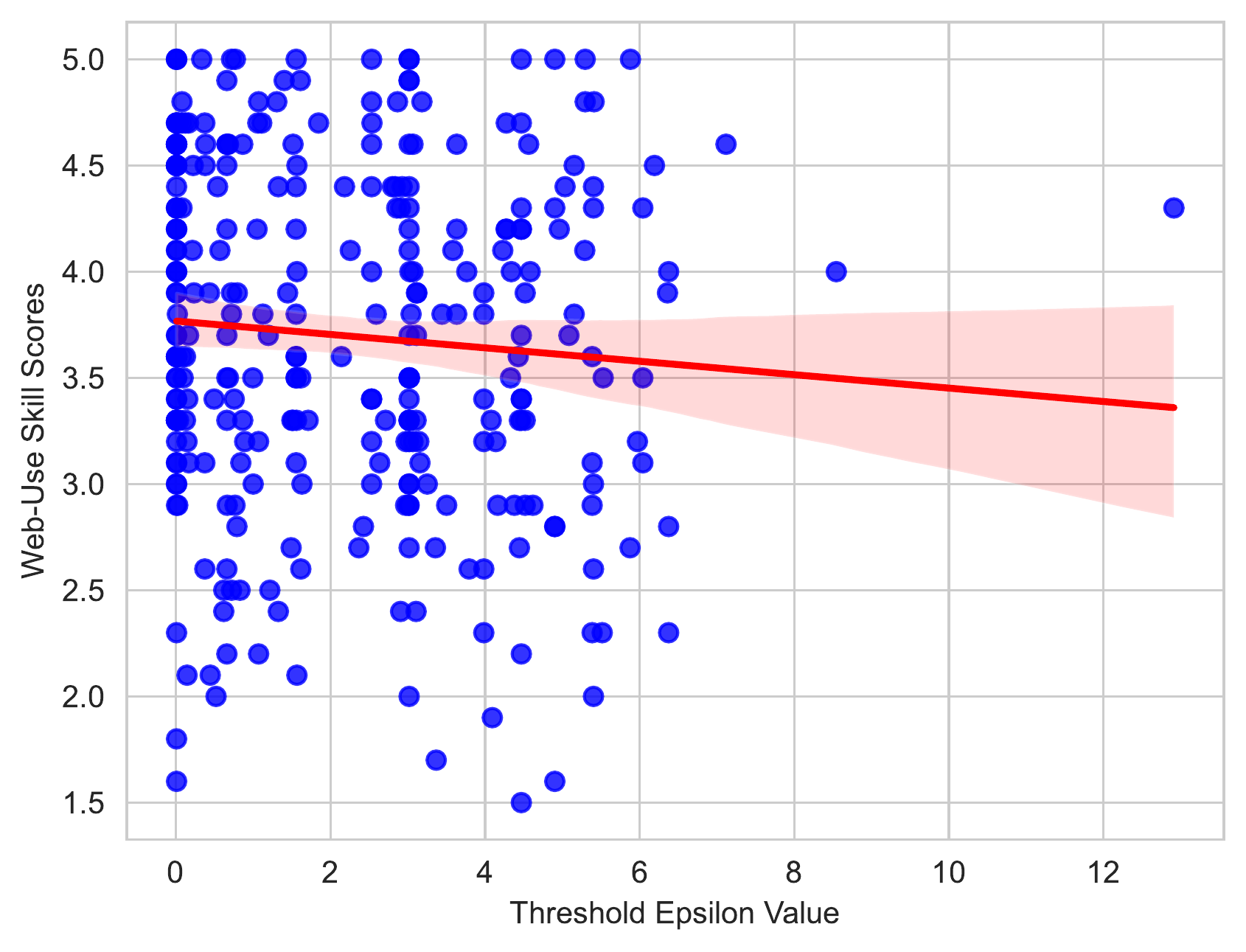}
	\caption{$\varepsilon$-Thresholds and \wus\ scores of all participants}
	\label{fig:05_EA_Behav_Exp_WUS_Thres}
\end{figure}

We fit a psychometric curve for each participant for each level of $n$ and determined the respective epsilon threshold. Based on $311$ participants times $5$ levels of $n$ this resulted in a total of $1,555$ epsilon thresholds. Thus we had five epsilon thresholds per participants, one per level of $n$. We computed the mean epsilon threshold based on these five values for each participant. We plotted these mean threshold values together with the \iuipc\ scores of all participants in \ref{fig:05_EA_Behav_Exp_IUIPC_Thres} and with the \wus\ scores of all participants in \ref{fig:05_EA_Behav_Exp_WUS_Thres}.

Figure \ref{fig:05_EA_Behav_Exp_IUIPC_Thres} shows Pearson correlation coefficient of $r=-0.16$, which is a significant negative linear relationship with $p=0.006$, however the correlation is not very strong. We conclude that the \iuipc questionnaire can be used as a means of determining privacy concerns but it is not a robust predictor of epsilon risks.
In contrast, Figure \ref{fig:05_EA_Behav_Exp_WUS_Thres} depicts the epsilon thresholds plotted against the \wus\ scores of the participants, which shows no significant relationship ($r=-0.08$, $p=0.162$). Based on this, we conclude that \wus\ is independent of perception of privacy risks.

\subsection{Discussion of a broader impact}

Let's assume we want to collect a dataset of size 10,000 samples originating from 10,000 distinct users. Such a dataset size is rather small for current deep-learning standards, however is not unrealistic for some specialized domains, say medical records. We would have to promise the data donors a system trained with $\varepsilon = 0.31$ on average which would most likely result in a useless system, given current techniques in model training with differentially privacy stochastic gradient descent \citep{senge-etal-2022-one}. Epsilon values in the range of 3.0--6.0 typical to current works would require a dataset of 10 million examples originating from 10 million distinct users. We are afraid that collecting such a dataset for building privacy-preserving systems is not realistic. Epsilon values over 10 impose such a privacy risk that nobody would be willing to share any data.

\section{Conclusion}

The research question of our study was whether there is a general and optimal epsilon value related to human risk assessments in the context of differential privacy. Through our survey and behavioral experiment, we found that the epsilon threshold was $\varepsilon_{\theta}=1.12$ across all conditions. However, further analysis revealed that the epsilon threshold was considerably impacted by the independent variables of ``\textit{Scenario}'' and ``\textit{Amount of Data Subjects}''. Based on our findings, we conclude that there is no general and optimal epsilon value, but rather that the epsilon threshold is dependent on the specific combination of the two independent variables. 

\section*{Limitations}

The limitations of our study should be considered when interpreting the results. Firstly, the sample size of $311$ participants may have limited the ability to generalize the findings to a larger population.
Second, there might be an inherent bias in the sample as mainly UK residents ended up in the pool of participants.
Furthermore, for each participant we have $225$ data points, i.e. ``\textit{Share}'' or ``\textit{Don't Share}'' responses. We had $45$ distinct stimuli combinations. Each combination got repeated five times resulting in the $225$ data points per participant. Both the number of data points per subject and the repetition rate of a single stimulus combination are relatively low. This was a conscious decision on our part, as the study was conducted online and therefore costs were incurred accordingly.

Especially in psychophysical experiments for threshold detection, it is not uncommon to collect several hundred, if not a thousand, data points per participant in order to have the corresponding threshold inferences more robust \citep{Kingdom_Prins_2016_Psychophysics}. In addition we applied a between-subject design with respect to the independent variable ``\textit{Scenario}''. To be able to compare the results of the distinct scenarios a within-subject design might lead to stronger conclusions.

\subsubsection*{Future work}

In future experiments, we would like to address some simplifications we made in this paper, namely the actual NLP scenarios. We only described them very coarsely (e.g., ``chat app", or "medical records''). However, it is worth exploring the setups deeper and find out what exactly in those documents made people worry about their privacy. At the same time, the incentives (sharing data for good) could be made more explicit (e.g., for what price).

\section*{Acknowledgments}

This project was partly supported by the PrivaLingo research grant (Hessisches Ministerium des Innern
und für Sport).

\bibliographystyle{lrec-coling2024-natbib}
\bibliography{bibliography}

\appendix


\section{Internet Users' Information Privacy Concerns (IUIPC)}
\label{sec:02_IUIPC}
The \textit{Internet Users' Information Privacy Concerns} (IUIPC) introduced by \citet{Malhotra_Kim_Agarwal_2004_Internet} refers to the concerns that individuals have regarding the collection, storage, use, and sharing of their personal information online.
Some specific examples of \iuipc\ include concerns about the sharing of personal information (individuals may be concerned about the sharing of their personal information with third parties without their knowledge or consent), or concerns about data retention (individuals may be concerned about how long companies retain their personal information).

\citet{Malhotra_Kim_Agarwal_2004_Internet} provide ten-item questionnaire to measure the attitude on privacy of users. Therefore \iuipc\ provides a way to understand and quantify the concerns that individuals have about their personal information online.
\citet{Vimalkumar_Sharma_2021_Okay} refer to the \iuipc\ construct when examining how voice-based digital assistants (VBDA) like Alexa, Siri and Google Assistants evoke serious privacy concerns regarding the collection, use and storage of personal data of the consumers. Their objective was to examine the perception of the consumers towards the privacy concerns and in turn its influence on the adoption of VBDA.
\citet{Rese_Ganster_Baier_2020_Chatbots} mention the \iuipc\ questionnaire in the context of the collection of personal data and the consideration of privacy concerns when evaluating chatbots and the acceptance of such chatbots by end-users


\section{Web-Use Skill}
\label{sec:02_WUS}
Web-use skill \citep{Hargittai_Hsieh_2012_Succinct} can be conceptualized as an individual's proficiency in utilizing the web for the purpose of information seeking and navigation. It encompasses a combination of technical and cognitive abilities, including the ability to effectively use search engines, evaluate the relevance and credibility of information, and utilize web-based tools and applications. Additionally, it includes the ability to protect oneself from online risks such as phishing and malware.

The research of \citet{Sindermann_Schmitt_2021_Online} is focused on the way in which the personal predispositions are associated with knowledge about how to protect one’s privacy online and actually protecting it. They investigated person characteristics underlying individual differences in online privacy literacy and behavior. \citet{Sindermann_Schmitt_2021_Online} use the \wus\ as part in determining the crystallized intelligence, introduced as a psychological construct by \citet{Cattell_1963_Theory}. Since \wus\ is an fundamental building block of crystallized intelligence according to \citet{Sindermann_Schmitt_2021_Online}, which in turn is used when examining online privacy literacy, it makes sense to add the \wus\ questionnaire to our study.

\section{Theoretical background}
\label{ch:02_Theoretical_Background}

This section summarizes the essential concepts for understanding differential privacy and various metrics for measuring personal risks, adapted from existing works.



\label{sec:02_Differential Privacy}
Differential privacy offers a formal treatment of protecting privacy of individuals whose data is being collected and analyzed \citep{Dwork_McSherry2006}. 
The fundamental hyperparameter of differential privacy is the privacy budget $\varepsilon$ which is proportional to the amount of information that an attacker can potentially learn about an individual from the released model or data. Smaller epsilon provides stronger privacy, but due to the need of adding more noise, the resulting analysis (e.g., the trained model) becomes poorer in terms of its accuracy.
The value of epsilon is usually chosen based on the desired level of privacy and the utility of the analysis. Typical values lay between $0.1$ and $10$, but it can be higher or lower depending on the application and the data \citep{Wood_Altman_2018_Differential}.




The most popular setup for using differential privacy is the `central' (also `global') differential privacy which means that data from individuals are held by a trusted curator who performs a private analysis, such as training a model \citep{Abadi_Chu_2016_Deep}.
Formally, having a random mechanism $M$, any two data sets $D_1$ and $D_2$ that differ in only one individual, and for any subset $S$ of possible outputs, differential privacy bounds the privacy loss by $\varepsilon$
\begin{equation}
	\frac{\Pr[M(D_1) \in S]}{\Pr[M(D_2) \in S]}	\leq \exp({\varepsilon})
\end{equation}
This means that the probability of any output of the mechanism $M$ on $D_1$ is at most $e^{\varepsilon}$ times the probability of the output of the same mechanism on $D_2$.

\subsection{Global privacy risk and local privacy leak}
\label{sec:02_GlobalPrvRisk_LocalPrivLeak}
In the context of differential privacy, the \textit{global privacy risk} refers to the risk that the data released or used for analysis reveals information about any individual in the data set beyond what can be inferred from the population as a whole \citep{Mehner_Voigt_Tschorsch_2021_Towards}. Global privacy risk is a measure of the privacy loss when a data set is released or used for analysis. It is related to the privacy budget $\varepsilon$ used in the definition of differential privacy.

The definition of the global privacy risk as well as the \textit{local privacy leak}, provided by \citet{Mehner_Voigt_Tschorsch_2021_Towards} is based on the work of \citet{Lee_Clifton_2011_How}, who developed a model that rephrases epsilon~($\varepsilon$) as a probability of re-identification depending on the number of data subjects in the data set and the sensitivity. They estimate the success probability of an adversary guessing the correct combination of data subjects in a data set based on the output of a differentially private mechanism. Further, they assume that the mechanism adds Laplace noise. \citet{Lee_Clifton_2011_How} define the privacy risk $\rho$, representing the probability of being identified as present or absent in a data set as data subject, as
\begin{equation}
	\rho \leq \frac{1}{1+(n-1) \exp{\left(-\varepsilon \frac{\Delta v}{\Delta f}\right)}}
\end{equation}

Here, $n$ represents the number of data subjects, $\Delta f$ is the sensitivity of the differentially private function, whereas $\Delta v$ represents the maximum change one of the data entries could cause on the function's result, therefore being the local sensitivity \citep{Lee_Clifton_2011_How}.

\paragraph{Running example} The following example is solely intended to simplify the illustration of the terminology and is taken from the work of \citet{Mehner_Voigt_Tschorsch_2021_Towards}.

Consider a school survey on drug abuse. To raise awareness, parents have access to the $\varepsilon$-differentially private results. Statistics such as the average age or the number of drug-using students per class can be obtained. Bob’s mother, Eve, finds out that there is high drug use in her son’s class. She wants to know who is using drugs. Eve queries the database for the average age of drug-addicted students of Bob’s class, which returns an $\varepsilon$-differentially private answer. Let us assume that the age of the students is between $0$ and $25$ years and that each class has at least one student who is recorded in the database. Hence, the sensitivity is given by $\Delta f = \frac{25-0}{1} = 25$, i.e., if Bob is $25$ years old, he would increase or decrease the average by $25$. However, students of the same class are typically the same age. For example, Eve knows that there are a total of $21$ students ($n = 21$) in Bob’s class, aged between $14$ and $18$. Additionally, with respect to the privacy risk $\rho$ defined by \citet{Lee_Clifton_2011_How}, we assume that only one student is not present in the database. Note that this corresponds to the worst case, since the number of combinations of possible students present is reduced to $n-1=20$. Accordingly, the local sensitivity yields $\Delta v = \frac{18-14}{20}=0.2$. Finally, assume the mechanism
uses $\varepsilon=\ln 3$. Hence, Eve’s probability of finding out which of Bob’s classmates use drugs yields $\rho \approx 5\%$.


\paragraph{Global sensitivity ratio} \citet{Mehner_Voigt_Tschorsch_2021_Towards} introduce the parameter $r$ which is defined as the \textit{global ratio of sensitivities}:
\begin{equation}
	r = \frac{\Delta v}{\Delta f}
\end{equation}

The privacy risk $\rho$ defined by \citet{Lee_Clifton_2011_How} therefore depends on the number of subjects $n$, the ratio of sensitivities $r$ and the privacy loss parameter~$\varepsilon$. According to \citet{Mehner_Voigt_Tschorsch_2021_Towards} the parameters $n$ and are $r$ are often unknown in advance, which makes it difficult to assess privacy risks. 

To overcome this limitation, \citet{Mehner_Voigt_Tschorsch_2021_Towards} propose the definition of the \textit{global sensitivity ratio}, which is based on a worst case assumption. Considering the worst case implies determining global values for $n$ and $r$. Remembering that $\Delta v$ is defined the maximum change one of the data entries could cause on the function's result, i.e. knowing that all students in Bob's class are aged between $14$ and $18$ as well as knowing the total number of students is $21$, and $\Delta f$ is the sensitivity of the function, i.e. knowing that all students in our example are aged between $0$ and $25$, the worst case would be that $\Delta v = \Delta f$, meaning there is Bob, a student that is actually $25$ years old. In this case, the maximum change caused by one of the present data subjects, i.e. a student increasing or decreasing the average age by $25$ years, corresponds to the sensitivity of the function, which is again based on the knowledge that all students in our example are aged between $0$ and $25$. Looking at the definition of $r$, the ratio of sensitivities, the following applies in the worst case: $r = \frac{\Delta v}{\Delta f}$, with $\Delta v = \Delta f$, therefore $r = \leq 1$ for all query functions.


\paragraph{Maximum Privacy Risk} Applying this to the privacy risk introduced by \citet{Lee_Clifton_2011_How}, \citet{Mehner_Voigt_Tschorsch_2021_Towards} introduce the \textit{maximum privacy risk} defined as:
\begin{equation}
	\rho \leq \frac{1}{1+(n-1) \exp(-\varepsilon)}
\end{equation}

In context of the running example provided by \citet{Mehner_Voigt_Tschorsch_2021_Towards}, this means that Bob’s age has the maximum possible impact on the mean, i.e., he is $25$ years old and the database contains only one person of his class. In this case, Eve would choose the correct present students and thus finding out who is using drugs with $13\%$ chance for $\varepsilon = ln 3$ and $n=21$\footnote{The original paper by \citet{Mehner_Voigt_Tschorsch_2021_Towards} states the a chance of $11$\%. However, inserting the values of $\varepsilon$ and $n$ into the function of the maximum privacy risk delivers a chance of $\rho \leq 13.04$\%}.


\paragraph{Global Number of Data Subjects} According to running example by \citet{Mehner_Voigt_Tschorsch_2021_Towards} we assume that only one student of Bob’s class is missing in the database. From this information alone, Eve can randomly guess which students are in the database. \citet{Mehner_Voigt_Tschorsch_2021_Towards} introduce $P_\textrm{guess}$ being the probability that an adversary can guess whether a data subject is present or absent in the data set. $P_\textrm{guess}$ is defined as:

\begin{equation}
	P_\textrm{guess} = \frac{1}{n}
\end{equation}

To go further in the direction of general values for $n$ and $r$, it makes sense looking at the impact of the number of data subjects $n$ with respect to the maximum privacy risk~$\rho$. Assuming the worst case of the data set just containing one data subject, i.e. $n = 1$, the maximum privacy risk is $\rho = 1$, independent of $\varepsilon$. This makes sense as an adversary will always choose the correct combination of data subjects if there is only one possible combination to choose from \citep{Mehner_Voigt_Tschorsch_2021_Towards}. It follows directly that $P_\textrm{guess} = 1$ and therefore, $\varepsilon$ has no influence in protecting the privacy of the data subject. Based on this the worst case, in which differential privacy still has an impact but success probabilities are maximized for an adversary, is given for $n = 2$.


\paragraph{Global Privacy Risk} Adding this worst case assumption of $n=2$ to the definition of the maximum privacy risk, the definition of the \textit{global privacy risk}~$P$ is given by \citet{Mehner_Voigt_Tschorsch_2021_Towards} as follows:

\begin{equation}
	P=\frac{1}{1+e^{-\varepsilon}}
\end{equation}

$P$ is the global upper bound of the maximum privacy risk $\rho$ with $n = 2$ and $r = 1$. Furthermore, $P_\textrm{guess} = \frac{1}{2}$. With increasing $\varepsilon$, the global privacy risk rises steadily and approaches $100\%$ without reaching it as can be seen in XXX provided by \citet{Mehner_Voigt_Tschorsch_2021_Towards}. Yet, a large $\varepsilon$ helps Eve to infer who is using
drugs in Bob's class with higher probability.



\paragraph{Global \& Local Privacy Leak} The privacy risk is an indicator of an adversary's success probability. However, according to \citet{Mehner_Voigt_Tschorsch_2021_Towards} it should be considered in the relation to $P_\textrm{guess}$ to determine the impact of a differentially private outcome on an adversary’s success probability. That is, the \textit{privacy leak} is the increment of the privacy risk caused by the release of an $\varepsilon$-differentially private result: $\Delta \rho = \rho - P_\textrm{guess}$.

\citet{Mehner_Voigt_Tschorsch_2021_Towards} claim that the privacy leak is not very intuitive since it is an increment to the guessing probability. Therefore it is suggested to consider the privacy leak as a relative increase by scaling it to a range from $0$ to $1$. Analogously to the global privacy risk, the maximum relative increase is given for $n = 2$ and $r = 1$. Based on this, \citet{Mehner_Voigt_Tschorsch_2021_Towards} introduce the \textit{global privacy leak}~$\Gamma$ as

\begin{equation}
	\Gamma = \frac{\Delta P}{1-P_\textrm{guess}}
\end{equation}

where $\Delta P = P - \frac{1}{2}$. Simultaneously, \citet{Mehner_Voigt_Tschorsch_2021_Towards} introduce the \textit{local privacy leak} $\gamma$ which is defined analogously with $\rho$ instead of $P$:

\begin{equation}
	\gamma = \frac{\Delta \rho}{1-P_\textrm{guess}}
\end{equation}

\end{document}